\documentclass[letterpaper, 10 pt, conference]{ieeeconf}  

\IEEEoverridecommandlockouts                              

\usepackage{graphicx}
\usepackage{subfig}
\usepackage[utf8]{inputenc}
\usepackage{tikz}
\usetikzlibrary{matrix,shapes,arrows,positioning,chains}
\usepackage{cite}

\title{\LARGE \bf
HMD Vision-based Teleoperating UGV and UAV for Hostile Environment using Deep Learning}

\author{Abhishek Sawarkar$^{1}$, Vishal Chaudhari, Rahul Chavan, Varun Zope, Akshay Budale and Faruk Kazi
\thanks{All authors are with Department of Electrical Engineering, Veermata Jijabai Technological Institute (VJTI), Mumbai, India.
        {$^{1}$\tt\small sawarkarabhi@gmail.com}}%
}

\begin{document}

\maketitle
\thispagestyle{empty}
\pagestyle{empty}

\begin{abstract}

The necessity of maintaining a robust anti-terrorist task force has become imperative in recent times with resurgence of rogue element in the society. A well equipped combat force warrants the safety and security of citizens and the integrity of the sovereign state. In this paper we propose a novel teleoperating robot which can play a major role in combat, rescue and reconnaissance missions by substantially reducing loss of human soldiers in such hostile environments. The proposed robotic solution consists of an unmanned ground vehicle equipped with an IP camera visual system broadcasting real-time video data to a remote cloud server. With the advancement in machine learning algorithms in the field of computer vision, we incorporate state of the art deep convolutional neural networks to identify and predict individuals with malevolent intent. The classification is performed on every frame of the video stream by the trained network in the cloud server. The predicted output of the network is overlaid on the video stream with specific colour marks and prediction percentage. Finally the data is resized into half-side by side format and streamed to the head mount display worn by the human controller which facilitates first person view of the scenario. The ground vehicle is also coupled with an unmanned aerial vehicle for aerial surveillance. The proposed scheme is an assistive system and the final decision evidently lies with the human handler.

\end{abstract}

\section{INTRODUCTION}

Governments around the world are investing billions of dollars to safeguard their citizen and secure their borders to maintain the territorial integrity and sovereignty of the nation. The risk of a terrorist attack can never be eliminated but prudent steps can be taken to reduce this risk. However, in certain cases due to the lapse in security measures, few rogue elements threaten to exploit human fears to help achieve their goals. They subjugate innocent civilians to get their demands fulfilled. With the advancements in technologies and advent of intelligent systems, these rogue elements can be neutralized with minimum human casualties. Intelligent robots can be used for combat or spy operations or for different purposes ranging from mine detection, surveillance, logistics and rescue operations to reconnaissance and support, communications infrastructure, forward-deployed offensive operations, and even as tactical decoys~\cite{sample1}. However, it is unreliable to give complete control to autonomous systems to handle such dangerous situations. Therefore, a semi-autonomous system in which computer intelligence assists the human operator will provide a better outcome. A system involving innate human intelligence coupled with robot's ability to process huge amounts of data can function effectively in such scenarios. The final call to make decision will evidently remain with the human while the machine could advocate its results based on the incoming data.
\par \indent Such systems are being developed by various defence agencies like DARPA's LAGR program~\cite{sample5} which focused on perception based off-road navigation in UGV's. Foster-Miller's TALON system is a remotely operated vehicle designed for missions ranging from combat to reconnaissance. They are working on incorporating virtual reality vision for their future vehicle called the  Modular Advanced Armed Robotic System (MAARS)\cite{sample6}. The US marines use their Gladiator Tactical Unmanned Ground Vehicle (TUGV) to minimize risks and eliminate threats during conflict. This vehicle is light weight and can be easily transported and deployed strategically for missions. The Indian Defence Research and Development Organization (DRDO) developed a fully automated UGV named Daksh for handling and destroying hazardous objects safely. These vehicles can be controlled remotely ensuring the safety of the operator and minimizing human loss.
\par \indent Tele-operation capabilities, or the ability for an operator to manipulate and control a robot remotely from a safe location through a radio link offers the possibility of controlling the robot in hostile environment with no human casualties. These capabilities reduce or remove operator's risk in highly stressful and dangerous environment. The human operator based on a remote location, away from the danger, receives the data wirelessly from the robot in real-time. This data is processed and presented upfront such that it does not overload the operator.
\indent This project proposes a robotic solution for handling terrorist situation in conjunction with human assistance. The machine can be used for first assault in hostile environment. The system comprises of tele-operated unmanned ground vehicle (UGV) and an unmanned aerial vehicle (UAV) that relays real-time information to the human controller wirelessly. The two robots provide multiple perspective of the environment to the controller \cite{sample2}. The UGV-UAV pair is inspired by the work done by Cantelli et al.\cite{sample4} for surveying operation. The system can classify terrorists from hostages by using a trained Deep Convolutional Neural Network. The control and operation of the machine is in the human controller's hands. By converting the video information in virtual reality (VR), the human controller can experience an immersive first person view of the situation\cite{sample3}. The person in consideration for classification is underscored by a bounding box with appropriate tag which is augmented on the video stream. The VR video system assists the operator to neutralize threat by detecting armed personnel in real-time.

\begin{figure*}[t]
\center
\includegraphics[width=0.8\textwidth]{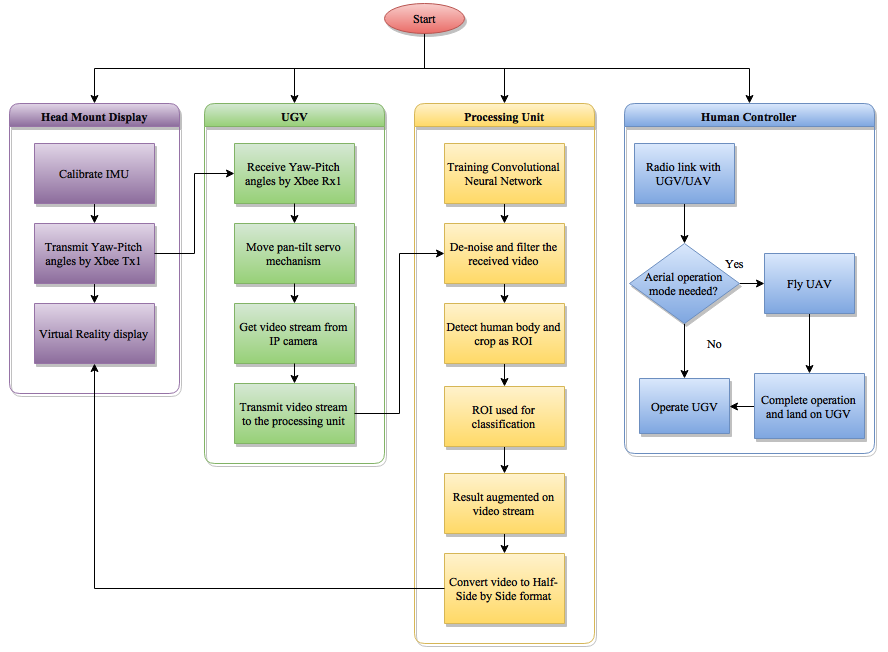}
\caption{Flowchart of proposed system}
\label{k}
\end{figure*}

\section{Hardware}
The hardware comprises of two robots, the Unmanned Ground Vehicle (UGV) for land based operation an the Unmanned Aerial Vehicle (UAV) for air surveillance.
\subsection{UGV}
An Unmanned Ground Vehicle (UGV) is a land based vehicle without a human on-board which can be used to perform civilian or dangerous military tasks. These vehicles are used to replace humans to work in perilous conditions like bomb defusal, dilapidated nuclear reactors, surveillance and tactical situations etc. 

The UGV used in this project has a six wheel differential drive mechanism that enables navigation in off-road and rugged terrain. The wheel mechanism is inspired by rocker-bogie system\cite{sample7} with additional wheels at the front and back of UGV. This mechanism facilitates the UGV to climb steps and steep slopes of maximum 50 degrees inclination. The UGV is equipped with a servo pan-tilt mechanism which holds an IP camera. The entire pan tilt mechanism is  mounted on the UGV base for stabilizing the video stream. This servo mechanism is controlled by head tracking inertial measurement unit(IMU) placed on the head mount VR goggle worn by the human controller. IP camera on the UGV wirelessly streams the reltime video input to a remote processing unit for classification and teleoperation.

\subsection{UAV}
The UAV stands for Unmanned Aerial Vehicle which is an aircraft without human on-board. UAV's can be remotely controlled by a human controller or it can fly autonomously based on pre-programmed flight plans. The UAV used in this project is remotely controlled first person view (FPV) type quad-rotor. This quad-rotor is mounted on the UGV base. Areas where the UGV fails to travel due to motion constraints, the operator can switch to aerial mode and fly UAV to complete the operation and land the UAV back on UGV. The Quadrotor frame is 'X' shaped of dimension 250mm measured diagonally from motor shaft-to-shaft. The carbon fiber body frame makes the UAV durable and light weight. A FPV camera relays the video stream when the remotely located human controller switches to the aerial mode. While the UGV could be used for combat scenarios providing attack capabilities in hostile environments, the UAV proposed in this project can only be used for reconnaissance and surveillance missions to provide vital information.

\begin{figure*}[!t]
\centering
\subfloat[UGV]{\includegraphics[width=2 in, height=2 in]{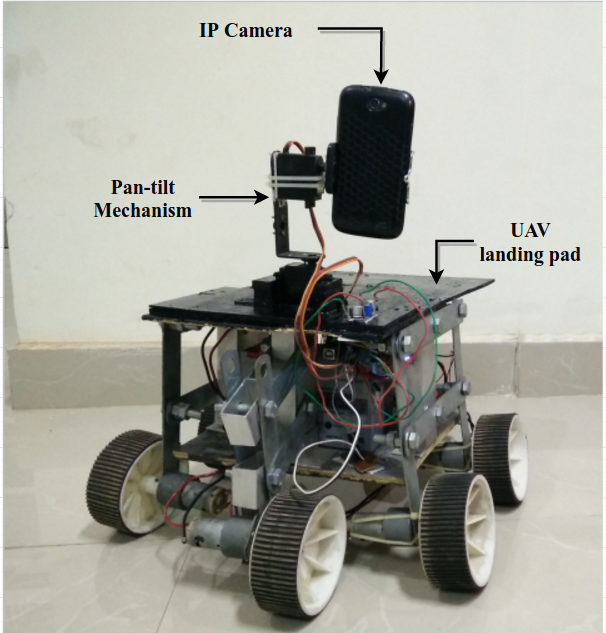}%
\label{fig_first_case}}
\hfil
\centering
\subfloat[UAV]{\includegraphics[width=2 in, height=2 in]{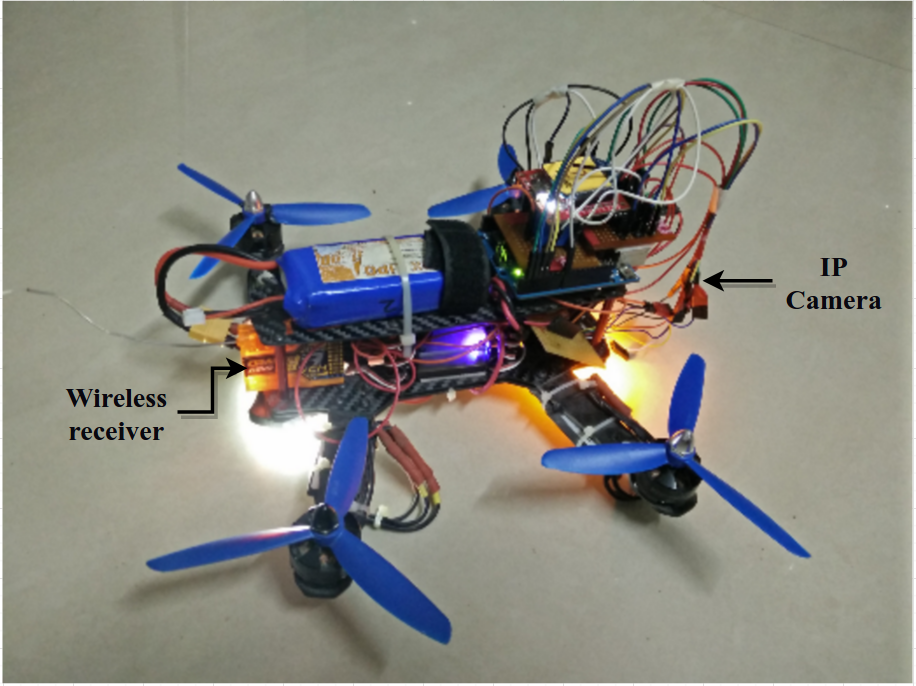}%
\label{fig_second_case}}
\caption{Unmanned Ground Vehicle and Unmanned Aerial Vehicle}
\label{fig_sim}
\end{figure*}

%

\section{Software}
\subsection{Image Processing}
The video stream received from the IP camera mounted on the UGV is pre-processed and analyzed by the remotely located processing unit. The video data is blurred to remove noise by Gaussian smoothing using a Gaussian kernel\cite{sample8}. Gaussian function transforms each pixel's original value to weighted average of that pixel's neighbourhood. The 2D Gaussian function is given by: 
\begin{equation}
G_{0}(x,y) = \frac{1}{2\pi\sigma^{2}}e^{(\frac{-(x-\mu_{x})^{2}}{2\sigma_{x}^{2}} + \frac{-(y-\mu_{y})^{2}}{2\sigma_{y}^{2}} )}
\end{equation}

$\mu_{x}$ = mean in horizontal axis
 
$\mu_{y}$ = mean in vertical axis

$\sigma^{2}$ = variance

\noindent After initial de-noising by the Gaussian filter, the video stream is processed to detect a human in the frame. Two algorithms are evaluated to find a person in the frame -

\subsubsection{HOG + SVM}

The Histogram of Oriented Gradients (HOG) descriptor for object recognition provides excellent performance for human detection. The local object appearance and shape can often be distinguished by the distribution of local intensity gradients or edge direction as described by Dalal et al.\cite{sample9}. The HOG descriptor identifies the feature set of the human figure based on the samples of positive and negative images provided during training. While testing the input images are classified as positive or negative by a linear Support Vector Machine (SVM). The frame containing a human figure is classified as positive else it is classified as negative. 

\begin{figure}[!h]
\centering
\includegraphics[width=2 in, height=1.5 in]{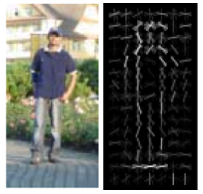}
\caption{Sample image of a person and its HoG visualization}
\label{ugv}
\end{figure}

The major disadvantage of the above method is that the classifier is used as a sliding window detector on an input image. The sliding window technique for classification of entire human frame is computationally expensive. Also, to detect all instances of humans at multiple scales, the sliding window has to run at multiple scales to form pyramid of detector responses. Any occlusion of the human results in loss of detection. The HOG+SVM algorithm used in this project was implemented using the OpenCV library to find a person in the video stream. The project demands a real-time performance as the robot will be used in hostile environment. However, the CPU implementation on Intel i3 with 4GB RAM and 2.4GHz processor was not realtime. An input video stream at 30 fps from IP camera on the UGV resulted in 8 fps output. The GPU implementation on Nvidia GPU GeForce GT 755M and 2GB RAM was better than the CPU with output at 20 fps. Another algorithm proposed by Viola and Jones\cite{sample10} was used to detect a face and the person in realtime.

\subsubsection{Haar feature based Cascade Classifier}
This algorithm incorporates different kind of feature set based on Haar wavelets instead of the usual image intensities as proposed by Papageorgiou et al. \cite{sample11}. A Haar-like feature stated in\cite{sample12} considers adjacent rectangular regions at a specific location in a detection window. It sums up the pixel intensities in each region and calculates the difference between these sums. This difference is then used to categorize subsections of an image. These Haar-like features can be calculated in constant time which make it faster than other algorithms. 

The main advantage of Haar based cascade classifier over other algorithms is the realtime detection of an object, in this case detection of a person. The algorithm proposed by Viola and Jones\cite{sample10} uses several simpler classifiers that are applied subsequently until at some stage the image is rejected or all stages are passed. If all stages are passed, the object of interest is detected and marked as the Region of interest (ROI). Based on human body geometry, the entire human anatomy is detected by the aforementioned algorithms and selected as the ROI. The ROI is cropped and resized to resolution of 227x227 pixels which is then forwarded to a trained Deep Network for classification of the person as probable terrorist.

\subsection{Deep learning and classification}
\begin{figure}[t]
\centering
\includegraphics[width=3.5 in, height=1.1 in]{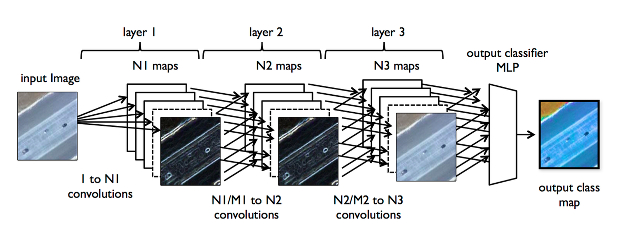}
\caption{General architecture of Convolutional Neural Network}
\label{ugv}
\end{figure}

A person detected as the ROI using human detection algorithm can be classified as a potential terrorist by assessing the features of a terrorist. These features may include weapons like a gun or assault rifle, facial emotions, walking patterns etc. This project considers a gun or assault rifle for classifying the person holding the weapon as a probable terrorist.

A Deep Learning neural network called the Convolutional Neural Network (CNN) is trained to identify human with guns like assault rifles, revolvers, etc. Deep learning aims at learning feature hierarchies wherein features in higher levels of the hierarchy are formed from lower level features extracted in previous stages\cite{sample13}. The salient advantage of CNN over other neural architectures is that the neurons in the higher layers have local connectivity with the neurons in the previous layer which reduces the number of connections and the weights to be trained. In order to improve generalization and reduce the number of hyper-parameters, a convolution operation on small region of the input image is performed. CNN utilizes 'shared weight' in the convolutional layers that reduces space complexity and improves performance.

The CNN architecture used in this project is based on Inception architecture\cite{sample14} formulated by GoogLeNet team which participated in ILSVRC14 competition which is based on the LeNet-5 network developed by LeCun et al.~\cite{sample15}. The network is 22 layers deep when counting only layers with parameters. The GoogLeNet team had trained the network on ImageNet dataset which comprises of 1.2 million images for training, 50,000 for validation and 100,000 for testing ranging across 1000 different categories. The final layer of the network is called the 'Softmax loss function' which is a linear layer that predicts the output out of 1000 classes. The 1000 different categories included both animate and inanimate objects. The network was trained to detect objects like car, hammer, gun, revolver etc for ImageNet Large Scale Visual Recognition Challenge 2015\cite{sample16}.

In this project, we utilize this pre-trained model to learn to classify particular object in the input image. We utilize 'Transfer Learning'\cite{sample17} technique in Caffe Deep Learning Framework\cite{sample18} to fine-tune the inception network in-order to detect probable terrorists as training the entire network for classification from scratch was not feasible given the limited data and hardware resources. The CNN was fine tuned to classify only 8 different classes instead of 1000 and the last layer of the network was changed in the model. The weights for the new last layer were randomly initialized and the model was re-trained on the new dataset comprising of different types of hand held weapons like assault rifle, revolver, pistol etc. The custom dataset constituted positive images of firearms and arbitrary negative samples of images not containing a weapon. Deep Convolutional Neural Network is capable of achieving better results for image classification than conventional algorithms. The network could achieve an error rate of 40\% on the custom terrorist dataset with the test inputs from highly noisy video stream. A bounding box of specific colour and prediction probability of the network are overlaid on the video stream to assist the human controller. 

If the human in the input is classified as a civilian, the bounding box is shown in 'Green' colour. However, if the network classifies the person as a probable terrorist, then the bounding box is displayed in 'Red'. Subsequently, the video stream is formatted and streamed to head mount display (VR goggles).

\begin{figure}[h]
\centering
\includegraphics[width=2.5 in, height=1.5 in]{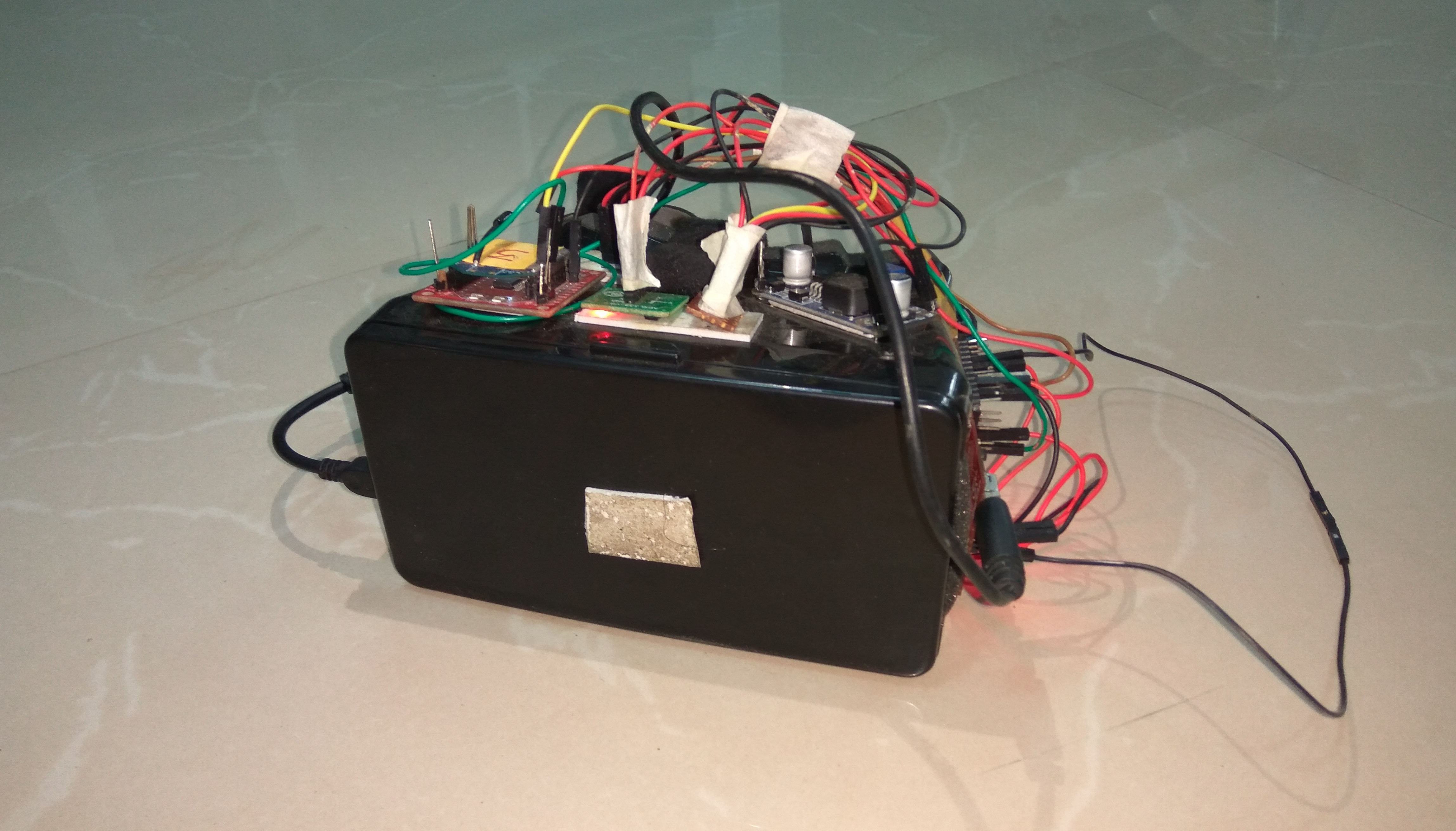}
\caption{Head mount display using VR goggle}
\label{ugv}
\end{figure}

\subsection{Head Mount Display}

Head mount displays (HMD) are increasingly being used in cockpits of modern helicopters and air-crafts to assist soldiers. These devices can display important tactical information along with the real scene that provides an effect of total immersion and a first person's perspective. The immersive vision helps the soldier to get a better understanding of the situation and control the remote robot with ease.

The HMD used in this project uses a smartphone virtual reality (VR) goggle that receives formatted video stream from the processing unit. The output frames given by the CNN are augmented with appropriate bounding boxes based on the classification results of CNN. These augmented frames are split into Half side-by-side format with each side having resolution of 950x1000. The frames are wirelessly streamed to the smartphone connected in the Ad-Hoc connection and placed in the VR google.

\begin{figure}[h]
\centering
\includegraphics[width=2.8 in]{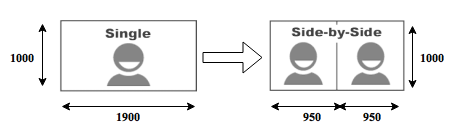}
\caption{Half Side-by-Side conversion}
\label{ugv}
\end{figure}

The human controller wearing the goggle gets a first person view of environment as seen by the UGV and UAV. The goggle is equipped with motion sensors (IMU) to track human controller's head movements which are calibrated at initialization. An accelerometer on the goggle records the pitch angle while a magnetometer notes the yaw orientation of the head movements. These angles are passed through a low pass filter to remove jitters and sharp reading changes. A Xbee transmitter relays the head tracking data to a Xbee receiver mounted on the UGV that controls the pan-tilt mechanism. As a result, the IP camera on the mechanism rotates as per the head movements of the human controller in real-time.

\section{Implementation}

The operator based at a remote location initiates the UGV via radio link. At the same time, the head mount display is calibrated by the operator to track his head movements. The head motion data is relayed to the UGV through XBee Rx-Tx pair based on IEEE 802.15.4 networking protocol. The UGV controls the motion of pan-tilt mechanism based on this data. Another encrypted radio link carries the visual input data of resolution 1900x1000 from an IP camera mounted on the pan-tilt mechanism on the UGV to the head mount display of the operator. Video data is pre-processed and before passing the visual data to the operator, a cloud server running a trained convolutional neural network scans whether the incoming data contains a probable terrorist or not. 

\subsection{Case I - Civilian}

\begin{figure}[h]
\centering
\includegraphics[width=1.6 in, height=1.2 in]{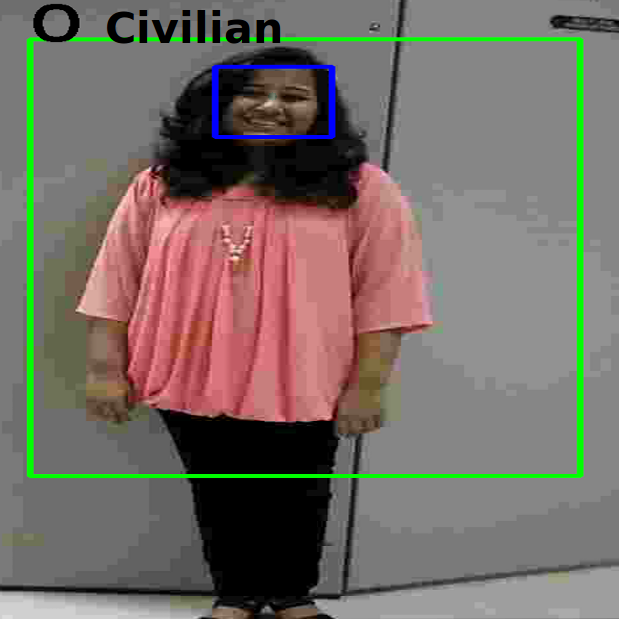}
\caption{As the detected human is unarmed, CNN classifies it as civilian and draws green bounding box around it. It also shows a confidence value (in this case 0), which is percent probability of the person being Terrorist.}
\label{ugv}
\end{figure}

\subsection{Case II - Armed personnel}
\begin{figure}[h]
\centering
\includegraphics[width=1.6 in,height=1.2 in]{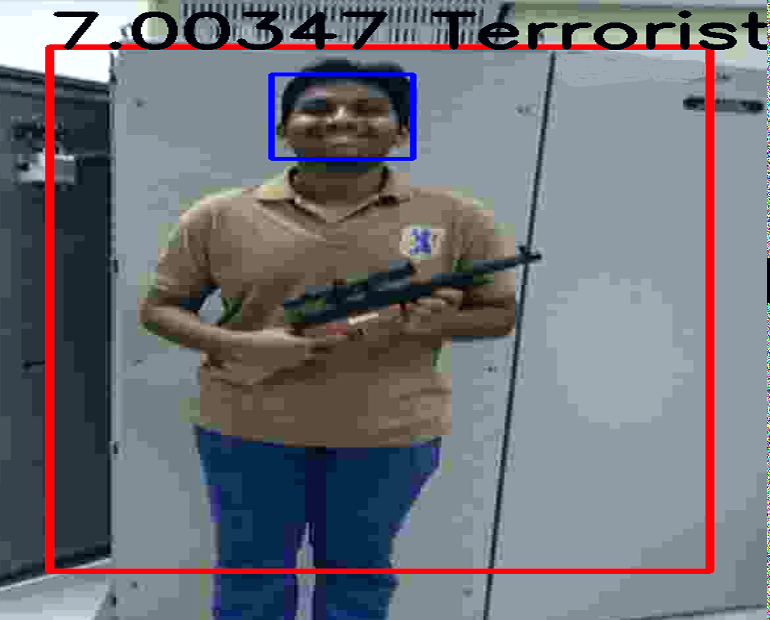}
\caption{As the detected human is armed, CNN classifies it as probable terrorist and draws red bounding box around it. The number indicates percent probability of the person being a terrorist}
\label{ugv}
\end{figure}

\noindent It classifies a person without a gun as a civilian with probability of being a terrorist close to null. A probable terrorist is a human with a gun in our model. The network presents it's classification with a confidence value as viewed at the top of the bounding box and stream is passed to VR formatter. The formatter resizes frames from resolution 1900x1000 into half-side by side configuration with each side of resolution 950x1000. With the head mount display, the operator gets an immersive real-time view of the environment seen by the robot. The situations wherein the UGV cannot operate or it's vision system is restricted, the operator can switch over to aerial mode and operate the UAV to get an aerial perspective of the situation.

\section{Conclusion and Future work}
In this paper we have presented a robotic solution for handling hostile situations thereby reducing casualties and loss of human life. The system comprises of both hardware and software entities to work in such environments. A pair of tele-operating UGV and UAV is proposed which is equipped with wireless vision sensor. Algorithms for human detection namely HOG+SVM and Haar Cascade are presented and a novel technique involving state of the art deep learning architecture for terrorist classification is exhibited for maintaining safety and security and use in hostile environment. The paper also underscores the advantages of head mount display in proposed tele-operating systems. 

\indent Currently the system identifies person with a gun as a threat. However, it can be designed to differentiate between an armed serviceman and an armed unknown person. This capability will eventually help in deploying the robot accompanied by task force to counter hostile situations. The face detection technique can be used to detect the person in consideration with pre-compiled database. A match will provide valuable information in such crisis.
  
\indent The system can be used in different scenarios like disaster response, rescue missions, operation in hazardous radioactive environment etc. The  deep convolutional network can be trained to identify specific entities like humans surviving in disaster hit areas, dangerous chemicals in decrepit chemical and radioactive plants, biological samples in such sites etc. A swarm of tele-operating robots, either fully autonomous or semi-autonomous could evidently prove beneficial in mitigating human loss in hostile situations.

\addtolength{\textheight}{-12.7cm}   




\section*{ACKNOWLEDGMENT}

The work presented in this paper was generously funded
by the Centre of Excellence (CoE) in Complex and Non-Linear
Dynamical Systems (CNDS) at VJTI.

\bibliographystyle{IEEEtran}
\bibliography{IEEEabrv,references}

\end{document}